\pdfoutput=1

\documentclass[11pt]{article}

\usepackage[final]{acl}

\usepackage{times}
\usepackage{latexsym}

\usepackage[T1]{fontenc}

\usepackage[utf8]{inputenc}

\usepackage{microtype}

\usepackage{inconsolata}

\usepackage{graphicx}
\usepackage{amsmath}
\usepackage{cleveref}
\usepackage{siunitx}
\usepackage{subcaption}
\usepackage{pdfcomment}
\usepackage{todonotes}

\newcommand{\pyannote}{\texttt{pyannote.audio} }
\newcommand{\wespeaker}{\texttt{wespeaker} }

\title{Playing with Voices: Tabletop Role-Playing Game Recordings as a Diarization Challenge}

\author{Lian Remme \\
  Heinrich Heine University Düsseldorf\\
  \texttt{lian.remme@uni-duesseldorf.de} \\\And
  Kevin Tang \\
  Heinrich Heine University Düsseldorf\\
  \texttt{kevin.tang@uni-duesseldorf.de} \\}

\usepackage{fancyhdr}
\fancypagestyle{firstpage}{
	\fancyhf{}
	\fancyfoot[L]{Copyright \copyright~2020 Association for Computational Linguists (ACL). All Rights Reserved.}

}

\begin{document}
\maketitle

\thispagestyle{firstpage}
\begin{abstract}
This paper provides a proof of concept that audio of tabletop role-playing games (TTRPG) could serve as a challenge for diarization systems. TTRPGs are carried out mostly by conversation. Participants often alter their voices to indicate that they are talking as a fictional character. Audio processing systems are susceptible to voice conversion with or without technological assistance. TTRPG present a conversational phenomenon in which voice conversion is an inherent characteristic for an immersive gaming experience. This could make it more challenging for diarizers to pick the real speaker and determine that impersonating is just that. We present the creation of a small TTRPG audio dataset and compare it against the AMI and the ICSI corpus. The performance of two diarizers, pyannote.audio and wespeaker, were evaluated. We observed that TTRPGs' properties result in a higher confusion rate for both diarizers.
Additionally, wespeaker strongly underestimates the number of speakers in the TTRPG audio files.
We propose TTRPG audio as a promising challenge for diarization systems.
\end{abstract}

\section{Introduction}

Speaker diarization is the process of determining how many people speak in a raw audio file and who spoke in which time frames~\cite{ryant21_interspeech,park2022review}. Rapid improvements have been made due to recent deep learning techniques~\cite{park2022review}.
However, the performance of diarization systems varies depending on the domain it is applied to, and is especially bad if multiple speakers talk in a restaurant setting~\cite{ryant21_interspeech}. 
It has also been shown that the performance is not yet good enough for in-person role-play dialogues in health care education~\cite{medaramitta2021evaluating}.

This study proposes tabletop role-playing games (TTRPGs) as a challenge for speaker diarization systems. TTRPGs are mostly played by conversation. Multiple people pretend to be characters and either describe their characters (\emph{descriptive}) or speak as their characters (\emph{in-character}). During in-character conversations, people usually alter voices, e.g. by adjusting tone or speed, or even by using an accent. This property of TTRPG makes it a natural challenge for diarization. A diarizer should be able to recognize which person is speaking, even if the speaker impersonates another (probably fictional) character. We provide a proof of concept for how TTRPGs can potentially be used as an additional benchmark for a diarizer. We create a small TTRPG audio dataset, apply \texttt{pyannote.audio}~\cite{Bredin23} (MIT license) and \texttt{wespeaker}~\cite{wang2023wespeaker} (Apache-2.0) on it and compare the diarizers' performance with their performance on the AMI dataset~\cite{kraaij2005ami} and the ICSI dataset~\cite{janin2003icsi} (both CC-BY 4.0). We show that the error rate, especially the confusion about who is speaking when, is higher for the TTRPG audio than other datasets. 
Furthermore, we find that \wespeaker underestimates the number of speaker in the TTRPG audio.

\section{Background}\label{background_section}
This section introduces diarization systems and how they can be fooled. We explain TTRPGs and why they could pose an interesting challenge, and we give details about the AMI and ICSI datasets.

\subsection{Challenges in diarization systems}\label{subsection:diarization_background}

There are various applications for diarization systems~\cite{nagavi2024comprehensive}, such as forensic analysis~\cite{grunert2023speaker}. Since the 1990s, there has been continuous development in diarization systems. New deep learning techniques have brought rapid improvements~\cite{park2022review}. Speech diarizers are usually trained on and/or evaluated against datasets like CALLHOME~\cite{CALLHOME}, the AMI corpus~\cite{kraaij2005ami}, the ICSI Meeting Corpus~\cite{janin2003icsi}, the dataset of the CHiME-6 challenge~\cite{watanabe2020chime}, People's Speech~\cite{galvez2021people}, or VoxConverse~\cite{Chung20}. These datasets are sourced from different speech domains and differ in a wide range of settings. They are unscripted telephone conversations~\cite{CALLHOME}, meeting recordings~\cite{kraaij2005ami,janin2003icsi}, dinner party recordings~\cite{watanabe2020chime}, or YouTube videos~\cite{Chung20}. Particularly notable diarizers include \texttt{pyannote.audio}~\cite{Bredin23,Plaquet23}, \texttt{wespeaker}~\cite{wang2023wespeaker} and \texttt{USTC-NELSLIP}~\cite{Wangetal2021system}. 

Identifying which domains or speaker behaviors are challenging for an audio processing tool is an important part of making models more robust in the future. In the most recent DIHARD Diarization Challenge \cite{ryant21_interspeech}, the best performing diarization system \cite{Wangetal2021system} achieved a Diarization Error Rate (DER) of $19.37\%$ in a domain-balanced evaluation set (core evaluation set) in Track 2 (diarization from scratch). Of the 11 domains examined, the three hardiest domains were speech in restaurant by 4 to 7 speakers, web videos mostly containing multiple speakers, and meetings containing 3 to 7 speakers with a DER ranging from $35\%$ to $45\%$ \cite{ryant21_interspeech}. This suggests that domains with three or more speakers which naturally contains more overlapping speech still pose a challenge for state-of-the-art systems.

The identification and mitigation of ways to worsen an audio processing result is an active area of research, especially in the field of speaker identification and diarization. One common approach in this field is voice spoofing, which is by creating a speech sample that mimics a target speaker (see \citet{yan2022survey} for an overview). Existing techniques include replaying speech samples recorded from a target speaker, speech synthesis, voice conversion, and human impersonation which is by mimicking a target speaker without technologies. Both voice professionals and non-professionals are able to spoof a system via impersonation, especially if the impersonator's natural voice is similar to that of the target speaker \cite{lau2004vulnerability,lau2005testing}.

Speaker recognition and diarization systems also struggle with non-adversarial attacks. For instance, speech from identical twins is a phenomenon which is difficult to recognize \cite{Revathi_forensic_2021} and distinguish because of similar vocal tract structure and other anatomical, physiological and physical characteristics. As even non-professionals are able to spoof speaker identification by changing their voices, speakers changing their voice to act as someone else could be a natural challenge for diarizers. This assumption is supported by the fact that diarizers' performances are not yet good enough for in-person role-play dialogues in health care education~\cite{medaramitta2021evaluating}.

\subsection{Tabletop Role-Playing Games -- TTRPGs}\label{section:background_dnd}

TTRPG are mostly played out by conversation. One or more players take the role of a character living in a world created by a game master (GM). Two types of conversations can occur: \textit{In-character} conversations, in which the participants talk as if they were characters, and \textit{descriptive} conversations, in which the participants say what their characters are doing or what is happening in the world. Most TTRPG consist of battle scenes in which the characters fight, and role-playing scenes in which the characters do something else, e.g. talk to each other or interact with the Non-Player Characters (NPCs). 

These conversations exhibit interesting linguistic properties. The linguistic information can falsely indicate a change of speaker. A sentence like “I walk towards the innkeeper: `Could I have something to drink?’”, could be said by one person. The role of the GM is particularly challenging, as they play every NPC in the world, and thus change between different direct speech without actually changing the speaker. For example, a GM says: “`Can I have something to drink?' --  `We have water' -- `One water, please!’” representing the speech of three characters. 

TTRPG players tend to change their voice during in-character conversations. This can be adjusting the speed rate, voice quality, pitch range or even speaking in an accent. This change can be a challenge for diarization systems~\cite{lau2004vulnerability,lau2005testing}. Even context switching can cause the diarization performance to drop, as lexical information contains information about when speakers changed~\cite{park2020speaker}.
The diarizer has access to lexical information in principle and could use it to determine speaker change. This could result into a performance drop when the lexical information about a speaker change is inaccurate. 

Due to these unique properties of TTRPG conversations, which we will be referring to as linguistic properties of TTRPGs, we look into whether TTRPGs can serve as a new speech domain for the evaluation of diarization systems.
A diarizer should be able to distinguish between a real change of speakers and the change of a voice due to impersonation. Only the former should be detected as change of speaker by a diarization system.
Evaluating diarization systems against TTRPG dialogues could lead to more robust diarizers.
Unintentional voice changes appear naturally in everyday conversations because of mood changes, quoted speech or to contextualize an utterance~\cite{gunthner1998polyphony}.
Therefore, robustness against voice change could be beneficial for diarizing of natural conversation as well.

\subsection{Reference datasets: AMI and ICSI corpora}\label{subsection:ami_dataset}

To evaluate how TTRPG properties influence diarization, we used reference datasets from a similar speech domain: %
The AMI corpus (headset mix)~\cite{kraaij2005ami} and the ICSI Meeting corpus (headset mix)~\cite{janin2003icsi,janin2004icsi}.
They share multiple properties with the TTRPG domain.
They are unscripted, conversational recordings from multiple speakers in a meeting scenario.
Both datasets are in English and contain native and non-native English speakers.
They have been annotated by multiple annotators and their agreement has been assessed.
However, the inter-annotator agreement has not been reported.

The AMI corpus consists of 170 audio files with \SI{97}{\hour} and \SI{40}{\minute} of audio, the ICSI corpus of 75 audio files with \SI{71}{\hour} and \SI{41}{\minute} of audio.

As one of the applied diarizers, \texttt{pyannote.audio}~\cite{Bredin23} (see \cref{ssec:diarizer}) was trained on the AMI corpus, it should perform particularly well on it. This means a comparison with the AMI corpus keeps our findings of \texttt{pyannote.audio} conservative. Therefore, we only used the test dataset which \pyannote was not trained on (16~files) from the AMI corpus on \texttt{pyannote.audio}.
These test files contain \SI{9}{\hour} and \SI{4}{\minute} of audio.

\section{Related Work}

Previous work suggested that TTRPG dialogues provide an appropriate challenge for artificial intelligence~\cite{ellis2017computers,martin2018dungeons}.
Datasets of \textit{written} TTRPGs conversations~\cite{callison2022dungeons,louis2018deep,rameshkumar2020storytelling} have been applied to different tasks, such as text generating~\cite{callison2022dungeons,newman2022generating,si2021telling} and character understanding~\cite{louis2018deep}.

We extend the application of TTRPG dialogues from the written to the audio domain. Audio processing is done, for example, in automatic speech recognition~\cite{huang2023mclf}, voice-based writing~\cite{goswami2023weakly}, or diarization~\cite{qamar2023speaking,qasemi2021paco} which we focus on.

Previous work showed that people can spoof speaker identification models by changing their voice~\cite{lau2004vulnerability,lau2005testing} and diarization models have poor performance for dialogues in which people pretend to be someone else~\cite{medaramitta2021evaluating}.
While TTRPG players usually do not try to mimick an existing person, many do change their voices while speaking in-character, thus posing a naturalistic challenge for diarization. %

\section{Data and methods}

This section provides an overview about how we created our TTRPG dataset and how we applied a diarizer on the audio files. Our TTRPG dataset was compiled using publicly available videos and subtitles. As this paper is a proof of concept and the data are not shared to any third-party, we had no ethical concerns in experimenting with the audio files (see \cref{subsection:ethics}). %
All scripts for the data processing and analyses, and links to the source videos we compiled the TTRPG data from are publicly available.\footnote{\url{https://github.com/LiRem101/playing-with-voices}}
The processed annotated transcripts are available from the authors upon request.

\subsection{Creating TTRPG dataset}\label{section:create_dataset}
The TTRPG dataset was created by extracting the audio files of English TTRPG campaigns from YouTube during August 2023. The campaigns was selected with the following steps: i) access YouTube in a Firefox browser (incognito mode), ii) search for “TTRPG campaign episode 1”, iii) include only videos that have manually added subtitles, and iv) select the first six campaigns with distinct (i.e., non-overlapping) speakers.
Only videos with manually added subtitles were used. This was done because speech-to-text systems do not perform well on natural conversations, and overlap of speech (which frequently happens in the videos) remains a key challenge~\cite{liesenfeld2023timing}.

Five of the six campaigns consisted of battle and role-playing scenes (see \cref{background_section}), one contained roleplay scenes only. We selected one hour of roleplay and, if available, one hour of battle audio from each campaign. We identified the onset of the roleplay portion using the first interaction between two player characters, and the onset of the battle portion when the GM signals the start of the battle. %
One campaign was removed, as cursory inspection revealed high subtitle inaccuracies. This resulted in 21~files with \SI{3}{\hour} and \SI{52}{\minute} of battle, and \SI{4}{\hour} and \SI{57}{\minute} of roleplay. %
The speakers speak American English.
We could determine the age of $72\%$ of the speakers (mean: $36$~years, min: $27$, max: $48$).
This was determined by checking the people's Wikipedia entries or their public social media channels and using the day the respective video was uploaded as the reference date.

The subtitles were used to create the ground truth diarization files. Some included the information about who spoke what, for others this was added manually. We applied forced alignment to the audio files and subtitles using \texttt{Wav2Vec2FABundle}~\cite{baevski2020wav2vec}. Combined with the knowledge of who said which words, we created diarization ground truth files by looking at the time stamps of each word and treating a gap of \SI{0.5}{\second} or less between two words of the same speaker as one utterance. While how short pauses are handled is not described for AMI and ICSI, a different approach from ours for the reference datasets should not influence our results, because we ignore \SI{0.5}{\second} around the start and the end of an utterance in the evaluation (see \cref{subsection:diar_result}).

A shortcoming of creating diarization ground truth this way is that overlapping speech as well as utterances that are not included in the subtitles files are not taken into account. These are further investigated in \cref{subsection:weaknesses_reference}. To overcome this, we manually annotated ten minutes of one role-playing audio manually without help of subtitle files or forced alignment. This annotation contains the information of whether the speech was in-character or not. As this is a proof of concept, the annotations was created by only one person (one of the authors, ANON) with the help of spectrograms, therefore inter-annotation-agreement was not evaluated.

\subsection{Applying the diarizers}\label{ssec:diarizer}

We applied the diarizers \texttt{pyannote.audio} (v. 3.1.0)~\cite{Bredin23} and \texttt{wespeaker} (v. 1.2.0)~\cite{wang2023wespeaker} on AMI, ISCI and our TTRPG dataset.
A computing cluster was used to diarize the audio files.
It took \SI{2.2}{\second} and \SI{0.67}{\second} per second of audio on average for \texttt{pyannote.audio} and \texttt{wespeaker} respectively, using a single CPU (Intel Xeon Gold 6136 (Skylake), \SI{3.00}{\giga\hertz}).

\subsubsection{\texttt{pyannote.audio}'s algorithm}

\texttt{pyannote.audio} %
diarizes audio files by first applying local speaker segmentation and embedding~\cite{pyannoteTechnical}.
This is done on overlapping frames of \SI{5}{\second} with \SI{500}{\milli\second} steps.
A maximum of three speakers can be determined in each frame.
Afterward, global agglomerative clustering is used to assign each local speaker to a global cluster.

\subsubsection{\texttt{wespeaker}'s algorithm}

\texttt{wespeaker} diarizes an audio file by first dividing the audio into frames of voice activity detected by Silero-VAD~\cite{sileroVAD}, and then applying a pre-trained ResNet34 architecture~\cite{zeinali2019but} on these frames for embedding extraction.

Spectral clustering is applied on the cosine similarity matrix of the embedding extraction to determine the number of speakers $n$.
The largest $n$ eigenvalues of the cosine similarity matrix and their corresponding eigenvectors are used to determine the active speaker(s) of each frame, by using kmeans and demanding $n$ clusters.
$n$ is either given, if the number of speaker is known, or determined by the largest difference between the $i$-th and $j$-th eigenvalue.
$i$ and $j$ default to $1$ and $20$.

\section{Results}

In this section, we present our findings about the diarization of TTRPG audio files compared to the AMI and ICSI dataset.
We used \texttt{pyannote.audio} (v. 3.1.0) and \texttt{wespeaker} (v. 1.2.0) to diarize 21 TTRPG audio files, the 171 (16 for \texttt{pyannote.audio}) AMI audio files, and the 75 ICSI audio files.
The evaluation was done using \texttt{pyannote.metrics} (v. 3.2.1)~\cite{pyannote.metrics}.
We applied Mann-Whitney U tests~\cite{mann1947test} to some result datasets to check whether the distributions of the datasets were significantly different, using a significance threshold of $5\%$.
In this section, we only comment if the differences were significant or not.
The exact values of the Mann-Whitney U tests can be found in \cref{sect:MWU-appendix}.

\subsection{Amount of detected speakers}\label{subsection:amount_speakers}

\begin{figure}[th]
  \centering
  \pdftooltip{\includegraphics[width=.9\linewidth]{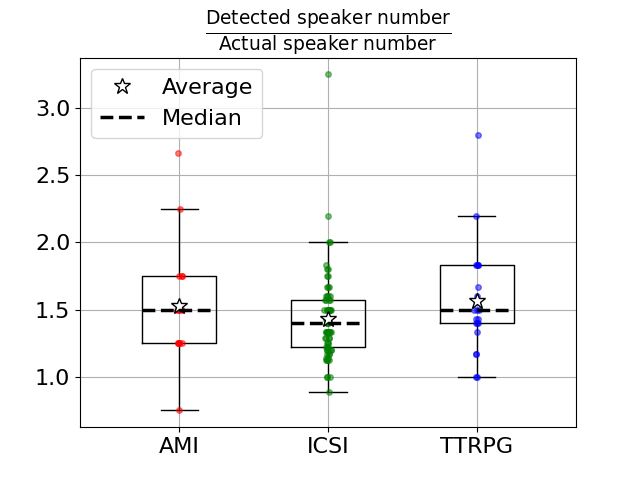}}{A boxplot diagram showing the detected speaker number divided through the actual speaker number for AMI, ICSI and TTRPG. AMI: Average 1.53, median 1.5, 1. quartile 1.25, 3. quartile 1.75, interquartile range from 0.75 to 2.25, outliners at 2.67. ICSI: Average 1.43, median 1.4, 1. quartile 1.22, 3. quartile 1.57, interquartile range from 0.89 to 2.0, outliners at 2.2 and 3.25. TTRPG: Average 1.56, median 1.5, 1. quartile 1.4, 3. quartile 1.83, interquartile range from 1 to 2.2, outliners at 2.8.}
  \caption{The relative amount of detected speakers by \texttt{pyannote.audio} compared to the amount of actual speakers in the audio files. Differences between TTRPG and AMI ($U=155$, $p=0.70$) and TTRPG and ICSI ($U=606.5$, $p=0.11$) were not significant (two-sided).}
  \label{fig:number_speakers_box}
\end{figure}

\begin{figure}[th]
  \centering
  \pdftooltip{\includegraphics[width=.9\linewidth]{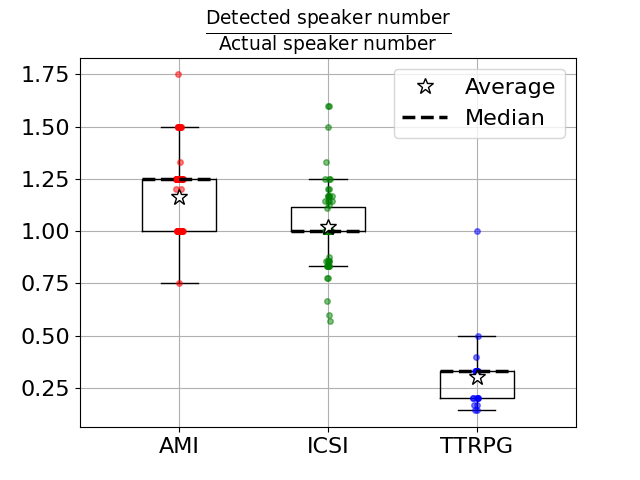}}{A boxplot diagram showing the detected speaker number divided through the actual speaker number for AMI, ICSI and TTRPG. AMI: Average 1.16, median 1.25, 1. quartile 1.00, 3. quartile 1.25, interquartile range from 0.75 to 1.50, outliner at 1.75. ICSI: Average 1.02, median 1.00, 1. quartile 1.00, 3. quartile 1.12, interquartile range from 0.83 to 1.25, outliners at 0.57, 0.60, 0.67, 0.78, 1.33, 1.50 and 1.60. TTRPG: Average 0.30, median 0.33, 1. quartile 0.2, 3. quartile 0.33, interquartile range from 0.14 to 0.50, outliner at 1.00.}
  \caption{The relative amount of detected speakers by \texttt{wespeaker} compared to the amount of actual speakers in the audio files. Significant differences (two-sided) were found between all distributions (AMI/ICSI $U=9459$, $p=2 \cdot 10^{-11}$, AMI/TTRPG $U=3495$, $p=2 \cdot 10^{-15}$, ICSI/TTRPG $U=1539$, $p=5 \cdot 10^{-12}$).}
  \label{fig:number_speakers_box_ws}
\end{figure}

In TTRPGs, the speakers change their voices during talking to emphasize that they talk in the role of a character. We therefore expected that a diarizer would detect more individual speakers in TTRPGs than dialogues that do not have this property, such as the AMI. 
The 21 TTRPG files contained $5.48$ speakers on average, the 171 AMI files $3.99$ (16 AMI files $3.94$) speakers on average, and the 75 ICSI files $5.95$ speakers on average.

The hypothesis of the diarizer finding more speakers in the TTRPG dataset was not supported.
We measured the relative amount of detected speakers divided by the amount of actual speakers.
The results of \texttt{pyannote.audio} can be seen in \cref{fig:number_speakers_box}.
It found  $1.52$ on average for the AMI dataset, $1.42$ for the ICSI dataset, and $1.56$ for the TTRPG dataset.
This difference was not statistically significant in a two-sided Mann-Whitney U test (\cref{tab:mann_whitney_u_amount_speakers}, \cref{sect:MWU-appendix}).
The results of \texttt{wespeaker} can be seen in \cref{fig:number_speakers_box_ws}.
It found $1.17$ on average for the AMI dataset, $1.01$ for the ICSI dataset, and $0.30$ for the TTRPG dataset.
A two-sided Mann-Whitney U test~\cite{mann1947test} showed significant differences between all distributions (\cref{tab:mann_whitney_u_amount_speakers}, \cref{sect:MWU-appendix}).

Considering AMI and ICSI, the relation of detected speakers against the actual number of speakers is significantly smaller for \wespeaker than for \texttt{pyannote.audio} (\cref{tab:mann_whitney_u_amount_speakers}, \cref{sect:MWU-appendix}), and \texttt{wespeaker}'s predictions are closer to the actual value.
However, \wespeaker drastically underestimates the number of speakers in TTRPG.

\subsection{Diarization errors}\label{subsection:diar_result}

The \emph{Diarization Error Rate} (DER) captures three types of errors~\cite{park2022review} -- \textit{Missed detection} (the diarizer failed to detect speech in the ground truth), \textit{False alarm} (the diarizer detected speech that is not in the ground truth) and \textit{Confusion} (the wrong speaker(s) assigned to detected speech). The DER is the sum of these errors divided by the audio time.
A \SI{0.5}{\second} collar around the start and the end of a speaker talking was removed from the evaluation (\SI{0.25}{\second} before and after, respectively).
This was done because it is difficult to pinpoint when exactly speech has begun.

All average error rates and Mann-Whitney U test results used in this section to check for significant differences can be found in \cref{tab:mann_whitney_u_diar_result} (\cref{sect:MWU-appendix}).

Both \texttt{pyannote.audio} and \texttt{wespeaker} had the highest average \textbf{DER} while diarizing the TTRPG, with $0.33$ (\texttt{pyannote.audio}) and $0.48$ (\texttt{wespeaker}) respectively (\cref{fig:der_box,fig:der_box_ws}, \cref{sec:fig-appendix}).
The results on the other datasets were significantly lower.
\texttt{pyannote.audio} reached an average of $0.12$ on AMI %
and $0.28$ on ICSI, while %
\texttt{wespeaker} had an average of $0.19$ on AMI %
and $0.27$ on ICSI (\cref{tab:mann_whitney_u_diar_result}, \cref{sect:MWU-appendix}). %

To understand this difference in DER better, we examined each of the three error types separately.

The \textbf{missed detection} rates between TTRPG and AMI are not significantly different for \pyannote %
or \texttt{wespeaker}%
, while ICSI differs significantly from AMI %
and TTRPG for both diarizers (\cref{tab:mann_whitney_u_diar_result}, \cref{sect:MWU-appendix}). %
The respective boxplots can be found in \cref{fig:missed_detection_box,fig:missed_detection_box_ws}, \cref{sec:fig-appendix}.

The TTRPG \textbf{false alarm} rates are significantly higher than the AMI rates for both \pyannote %
and \texttt{wespeaker}. %
The TTRPG dataset's false alarm also differs significantly from the ICSI dataset, %
 but is not significantly higher, since the ICSI dataset shows the highest false alarm rate of all datasets (\cref{tab:mann_whitney_u_diar_result}, \cref{sect:MWU-appendix}).
The distributions can be seen in \cref{fig:false_alarm_box,fig:false_alarm_box_ws}, \cref{sec:fig-appendix}.

As with both diarizers ICSI has higher false alarm but lower missed detection than AMI, we suspect that the differences between those two datasets root in a more careful annotation on ICSI's reference files.
This is backed up by fewer interjections in the ICSI reference files than in the AMI reference files (see \cref{subsection:weaknesses_reference}) and the fact that both diarizers showed this result.

However, the diarizers do not score well on missed detection or false alarm on the TTRPG dataset.
We suspected that this was not due to the poor performance of the diarizer, but by the quality of the ground truth files. This is further evaluated in \cref{subsection:weaknesses_reference,subsection:manual_labelled_eval}.

\begin{figure}[t]
  \centering
  \pdftooltip{\includegraphics[width=.9\linewidth]{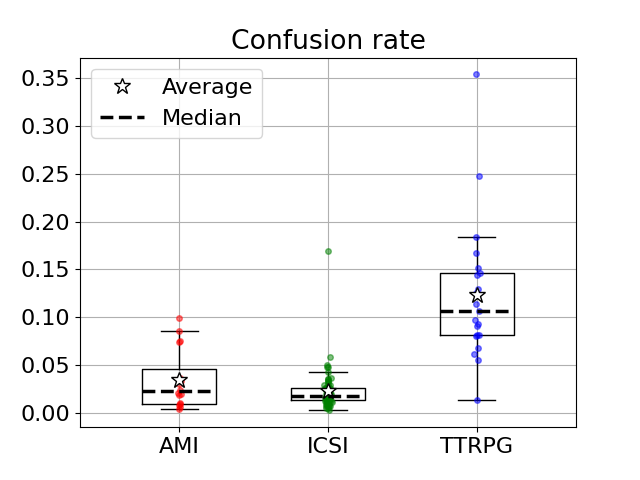}}{A boxplot diagram showing the confusion rate for AMI, ICSI and TTRPG. AMI: Average 0.034, median 0.022, 1. quartile 0.01, 3. quartile 0.045, interquartile range from 0.004 to 0.085, outliners at 0.1. ICSI: Average 0.022, median 0.017, 1. quartile 0.013, 3. quartile 0.025, interquartile range from 0.002 to 0.043, outliners at 0.048, 0.048, 0.050, 0.058 and 0.169. TTRPG: Average 0.123, median 0.106, 1. quartile 0.081, 3. quartile 0.145, interquartile range from 0.013 to 0.183, outliners at 0.25 and 0.35.}
  \caption{Confusion rates of the AMI, ICSI and the TTRPG datasets by \texttt{pyannote.audio}. TTRPG shows significantly higher confusion than AMI ($U=32$, $p=2 \cdot 10^{-5}$, one-sided test) and ICSI ($U=75$, $p=1 \cdot 10^{-10}$, one-sided test).}
  \label{fig:confusion_box}
\end{figure}

\begin{figure}[t]
  \centering
  \pdftooltip{\includegraphics[width=.9\linewidth]{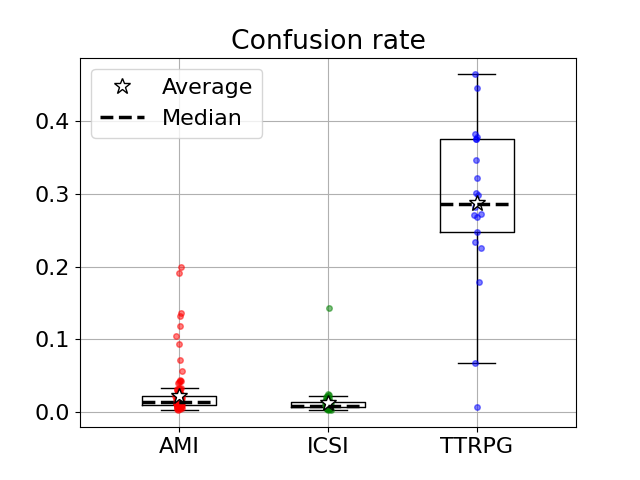}}{A boxplot diagram showing the confusion rate for AMI, ICSI and TTRPG. AMI: Average 0.022, median 0.015, 1. quartile 0.011, 3. quartile 0.022, interquartile range from 0.003 to 0.040, outliners at 0.042, 0.043, 0.044, 0.045, 0.057, 0.072, 0.093, 0.105, 0.118, 0.132, 0.136, 0.192 and 0.199. ICSI: Average 0.012, median 0.009, 1. quartile 0.007, 3. quartile 0.013, interquartile range from 0.002 to 0.023, outliner at 0.143. TTRPG: Average 0.287, median 0.286, 1. quartile 0.247, 3. quartile 0.375, interquartile range from 0.067 to 0.464, outliner at 0.007.}
  \caption{Confusion rates of the AMI, ICSI and the TTRPG datasets by \texttt{wespeaker}. TTRPG shows significantly higher confusion than AMI ($U=165$, $p=7 \cdot 10^{-12}$, one-sided test) and ICSI ($U=54$, $p=4 \cdot 10^{-11}$, one-sided test).}
  \label{fig:confusion_box_ws}
\end{figure}

\Cref{fig:confusion_box,fig:confusion_box_ws} show the confusion of the diarizers for the TTRPG dataset compared to the AMI and ICSI dataset.
For the \textbf{confusion}, the TTRPG rates are significantly higher than the AMI rates %
and the ICSI rates %
for both diarizers, while the confusion of AMI and ICSI datasets does not differ significantly on \texttt{pyannote.audio}%
, but does differ on \wespeaker (\cref{tab:mann_whitney_u_diar_result}, appendix).%

To establish whether these performance differences are independent of the diarizer's ability to detect the number of speakers (\cref{subsection:amount_speakers}), we repeated our diarization experiments with the actual number of speakers given to the system. This modification did not influence the error rates of missed detection and false alarm. While we expected that it would decrease the confusion error, it in fact led to an increase of the AMI and ICSI dataset confusion on \texttt{pyannote.audio}, to a point where the difference between both datasets was no longer significant (AMI: $0.034 \rightarrow 0.19$, $+459\%$, ICSI: $0.022 \rightarrow 0.15$, $+582\%$, TTRPG: $0.12 \rightarrow 0.15$, $+25\%$). %
On \texttt{wespeaker}, the average confusion rates changed for AMI and ICSI if we gave the number of speakers to the diarizer (AMI: $0.022 \rightarrow 0.018$, $-19\%$, ICSI: $0.012 \rightarrow 0.019$, $+58\%$, TTRPG: $0.29 \rightarrow 0.29$, $\pm 0\%$).
However, the AMI and ICSI confusion stayed significantly smaller than the TTRPG confusion (\cref{tab:mann_whitney_u_diar_result}, \cref{sect:MWU-appendix}). %
The respective boxplots can be found at \cref{fig:confusion_speaker_box,fig:confusion_box_speaker_ws}, \cref{sec:fig-appendix}.

\subsection{Weaknesses of the TTRPG reference files}\label{subsection:weaknesses_reference}

The TRPG reference files were created with subtitles and forced alignment, as described in \cref{section:create_dataset}.
While with this approach the ground truth can be efficiently obtained, it does negatively influence its quality because it depends on the quality of the subtitles and of the forced alignment process. This has been especially evident when we had to exclude one of the chosen campaigns, after noticing its gross error rates and finding that its subtitle was far from being verbatim (see \cref{section:create_dataset}).

Even after excluding campaigns with gross transcription issues, the approach of using subtitles had two problems.
First, subtitles often lacked filler words like “uhmm” or “ehh” and paraphrased the meaning of words. Secondly, the use of subtitles and forced alignment ignored overlapping speech by multiple speakers. It either ignores all but one speaker or appears as if the words would have been said after each other. Both problems would lead to missing speech in the ground truth, which can result in an apparently higher false alarm rate we observed in \cref{subsection:diar_result} compared to AMI. To verify this further, we evaluated the properties of the TTRPG dataset and compared it to AMI and ICSI.

spaCy's \texttt{en\_core\_web\_trf} (v. 3.7.3)~\cite{spacy2} was used to identify the number of words tagged as an interjection in our ground truth, AMI, and ICSI. The TTRPG dataset subtitles have $4\%$ of interjections, while AMI transcriptions have $13\%$ and ICSI transcriptions $9.5\%$. Additionally, we calculated the amount of overlapped speech given by the reference files. The TTRPG files consisted of $0.2\%$ of overlapped speech, while AMI has $6.5\%$ and ICSI $3.8\%$. These differences suggest that the reference files of our own dataset do not depict overlapping speech sequences correctly, and that some utterances may be missing.
This is also backed by the manually annotated reference TTRPG file, which contains $4,6\%$ of overlapped speech.

\subsection{Manually annotated reference file}\label{subsection:manual_labelled_eval}

To estimate how much the aforementioned issue of the reference files influenced our results, we annotated \SI{10}{\minute} of one TTRPG audio file manually with the information which speaker spoke at what time and repeated and extended our error analyses.\footnote{To manually annotate the whole TTRPG dataset was not feasible given our resources and we will leave this for future research.
} %
To get a representative example of TTRPG, we chose a role-playing file whose DER, confusion, false alarm and missed detection rate was among the ones closest to the average. %
We annotated a continuous portion that contained frequent speech from all speakers and in-character conversations. 

The resulting error rates can be seen in \cref{tab:error_manual_annotation}, compared to the error rates from the automatically created reference files.
The manually created reference files result in overall smaller error rates when \pyannote was used. While the confusion rate only decreased by $5\%$, the false alarm decreased by $61\%$ and the missed detection by $50\%$. If the number of speakers was given to the diarizer upfront, the rates decreased by $8\%$, $60\%$, and $40\%$ respectively.
For \texttt{wespeaker}, all error rates but missed detection got smaller. The confusion rate decreased by $13\%$ ($8\%$ if speaker number was given) and the false alarm by $56\%$ ($71\%$). The missed detection rate raised by $83\%$ ($83\%$).

These results suggest that the higher confusion error on TTRPG datasets was due to the linguistic properties of TTRPG datasets, while the higher false alarm rates were caused by imperfect reference files.\footnote{We cannot offer an explanation for why the missed detection for wespeaker increases. The point that the quality of the reference files does not influence our findings is not affected.}

\setlength{\tabcolsep}{3pt}
\begin{table}
  \centering
  \begin{tabular}{ c c c c c c c }
    \hline
    \begin{tabular}{@{}c@{}}\textbf{Anno-}\\\textbf{tation}\end{tabular} & \begin{tabular}{@{}c@{}}\textbf{Diar-}\\\textbf{izer}\end{tabular} & \begin{tabular}{@{}c@{}}\textbf{Speaker}\\\textbf{num.}\\\textbf{given}\end{tabular} & \textbf{DER} & \textbf{Conf} & \textbf{FA} & \textbf{MD}\\
    \hline
    MAN & PA & No & $0.30$ & $0.20$ & $0.07$ & $0.03$\\
    AUTO & PA & No & $0.45$ & $0.21$ & $0.18$ & $0.06$\\
    \hline
    MAN & PA & Yes & $0.35$ & $0.25$ & $0.08$ & $0.03$\\
    AUTO & PA  & Yes & $0.52$ & $0.27$ & $0.20$ & $0.05$\\
    \hline
    MAN & WS & No & $0.55$ & $0.40$ & $0.04$ & $0.11$\\
    AUTO & WS & No & $0.61$ & $0.46$ & $0.09$ & $0.06$\\
    \hline
    MAN & WS & Yes & $0.47$ & $0.34$ & $0.02$ & $0.11$\\
    AUTO & WS  & Yes & $0.50$ & $0.37$ & $0.07$ & $0.06$\\
    \hline
  \end{tabular}
  \caption{
    The error rates created by 10 min of manually annotated reference files compared to the error rates that are created by the subtitle reference files.MAN: manual, AUTO: automatic, DER: diarization error rate, Conf: confusion; FA: false alarm, and MD: missed detection. PA: \texttt{pyannote.audio}, WS: \texttt{wespeaker}.
  }\label{tab:error_manual_annotation}
\end{table}

\subsection{Detailed error analyses}

\begin{figure*}[ht]
	\begin{subfigure}{0.45\textwidth}
        \pdftooltip{\includegraphics[height=5.7cm]{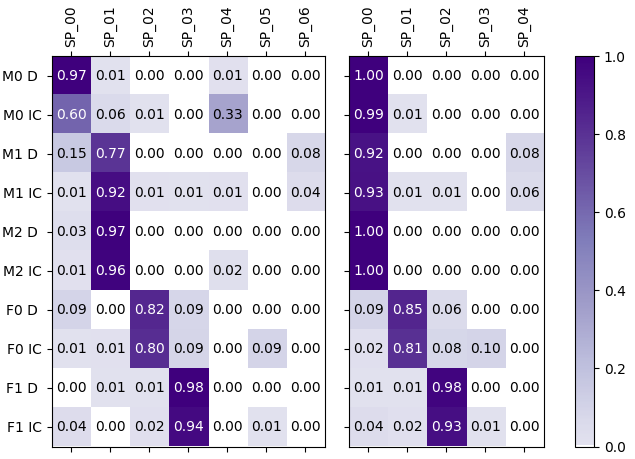}}{Two confusion matrices. Left one: The x-axis has the values SP\_00, SP\_01, SP\_02, SP\_03, SP\_04, SP\_05 and SP\_06. The y-axis has the values M0 D, M0 IC, M1 D, M1 IC, M2 D, M2 IC, F0 D, F0 IC, F1 D and F1 IC. The values are [[0.97, 0.01, 0.00, 0.00, 0.01, 0.00, 0.00], [0.60, 0.06, 0.01, 0.00, 0.33, 0.00, 0.00], [0.15, 0.77, 0.00, 0.00, 0.00, 0.00, 0.08], [0.01, 0.92, 0.01, 0.01, 0.01, 0.00, 0.04], [0.03, 0.97, 0.00, 0.00, 0.00, 0.00, 0.00], [0.01, 0.96, 0.00, 0.00, 0.02, 0.00, 0.00], [0.09, 0.00, 0.82, 0.09, 0.00, 0.00, 0.00], [0.01, 0.01, 0.80, 0.09, 0.00, 0.09, 0.00], [0.00, 0.01, 0.01, 0.98, 0.00, 0.00, 0.00], [0.04, 0.00, 0.02, 0.94, 0.00, 0.01, 0.00]]. Right: The x-axis has the values SP\_00, SP\_01, SP\_02, SP\_03 and SP\_04. The y-axis has the values M0 D, M0 IC, M1 D, M1 IC, M2 D, M2 IC, F0 D, F0 IC, F1 D and F1 IC. The values are [[1.00, 0.00, 0.00, 0.00, 0.00], [0.99, 0.01, 0.00, 0.00, 0.00], [0.92, 0.00, 0.00, 0.00, 0.08], [0.93, 0.01, 0.01, 0.00, 0.06], [1.00, 0.00, 0.00, 0.00, 0.00], [1.00, 0.00, 0.00, 0.00, 0.00], [0.09, 0.85, 0.06, 0.00, 0.00], [0.02, 0.81, 0.08, 0.10, 0.00], [0.01, 0.01, 0.98, 0.00, 0.00], [0.04, 0.02, 0.93, 0.01, 0.00]].}
		\subcaption{Confusion matrix created by \texttt{pyannote.audio}, right if the speaker number is given, left if it is not.}%
		\label{fig:pyannote-confusion}
	\end{subfigure}%
    \hspace{0.1\textwidth}
    \begin{subfigure}{0.45\textwidth}
		\centering
        \pdftooltip{\includegraphics[height=5.7cm]{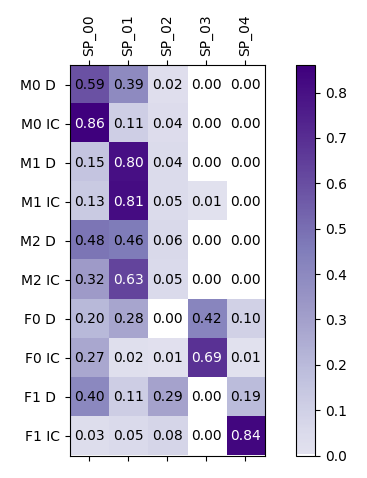}}{A confusion matrix. The x-axis has the values SP\_00, SP\_01, SP\_02, SP\_03 and SP\_04. The y-axis has the values M0 D, M0 IC, M1 D, M1 IC, M2 D, M2 IC, F0 D, F0 IC, F1 D and F1 IC. The values are [[0.59, 0.39, 0.02, 0.00, 0.00], [0.86, 0.11, 0.04, 0.00, 0.00], [0.15, 0.80, 0.04, 0.00, 0.00], [0.13, 0.81, 0.05, 0.01, 0.00], [0.48, 0.46, 0.06, 0.00, 0.00], [0.32, 0.63, 0.05, 0.00, 0.00], [0.20, 0.28, 0.00, 0.42, 0.10], [0.27, 0.02, 0.01, 0.69, 0.01], [0.40, 0.11, 0.29, 0.00, 0.19], [0.03, 0.05, 0.08, 0.00, 0.84]].}
		\subcaption{Confusion matrix for \wespeaker (given speaker number). No given number resulted in one detected speaker.}%
		\label{fig:wespeaker-confusion}
	\end{subfigure}%
	
    \caption{The confusion matrices for the \SI{10}{\minute} manually annotated audio. The audio contained 5 speakers, 3 male (M0 to M2) and 2 female (F0 and F1). The reference (rows) for each speaker is divided into what the speaker said in their descriptive voice (D) and their in-character voice (IC). The predictions are given by the columns, normalized with respect to the reference. Shown for \pyannote result (\cref{fig:pyannote-confusion}) and \wespeaker result (\cref{fig:wespeaker-confusion}).}
    \label{fig:confusion_matrices}
\end{figure*}

To further understand the errors of the diarizer, we examined the confusion matrices for the manually created reference (see \cref{fig:confusion_matrices}; the rows of the matrices are normalized).
For the matrices, we only examined regions that have been labeled as speech in the reference and the prediction, ignoring false alarm and missed detection. The manually created reference contained information whether speech was “in-character” or “descriptive”. We differentiated in the matrices between these two ways of speaking for every person. 

\Cref{fig:pyannote-confusion} (left) is the confusion matrix when \pyannote has not been given the number of actual speakers.
Two male speakers (M1 and M2) have been “merged” and are mostly mapped to the predicted SP\_01. In this audio file, M0 has had the position of the GM, meaning that the speaker had not one in-character voice but several (see \cref{section:background_dnd}). This seems to lead to wrong predictions by the diarizer, such as that in-character of M0 “creates” a new speaker SP\_04. Additionally, SP\_0 has been assigned to many other actual speakers, especially M1 and F0.

\pyannote identifies three speakers that get mapped to small amounts of actual speakers. SP\_04 is a speaker created by M0's in-character voice.
SP\_05 is a speaker created mostly by the in-character voice of F0. SP\_06 seem to be parts of the actual M1.
It is interesting to note that the in-character voices of the two female voices (F0~IC and F1~IC) and M0~IC seem to be more distinctive “speakers” respectively than M1 and M2, who are merged into one detected speaker. To examine how distributed the mappings are,  Shannon entropy \cite{Shannon1948} was calculated over each row in the confusion matrix (the higher the entropy, the more distributed a mapping is). The average entropy over all rows, descriptive voices and in-character voices are $0.59$, $0.48$ and $0.70$ respectively.
This shows that the change of voice of the players results into a higher scattering of detected speakers if \pyannote is used.

\Cref{fig:pyannote-confusion} (right) is the confusion matrix when \pyannote has been given the number of actual speakers.
All male speakers are mapped to the same predicted speaker. Two extra speakers (SP\_04 and SP\_03) consist of some of M1's utterances and the in-character voice of F0. Again, the change of voice leads to a higher scattering of detected speakers. The average entropy over all rows, descriptive voices and in-character voices are $0.32$, $0.27$, and $0.38$ respectively.

\Cref{fig:wespeaker-confusion} shows the confusion matrix when \wespeaker has been given the number of actual speakers.
A confusion matrix without the speaker number given has not been created, since \wespeaker predicted only one speaker in this case.
This aligns with our findings that \wespeaker finds small numbers of speakers for the TTRPG files (see \cref{subsection:amount_speakers}).
The average entropy over all rows, descriptive voices and in-character voices are $1.17$, $1.38$, and $0.96$.
In the case of \texttt{wespeaker}, the in-character voices are not more scattered than the descriptive voices.
However, the scattering is higher than \texttt{pyannote.audio}'s overall.

\section{Conclusion}
This paper aimed to provide a proof of concept that TTRPG audio files serve as a complex but natural challenge to diarization systems.
We were able to show that \texttt{pyannote.audio}'s and \texttt{wespeaker}'s speaker confusion increased for TTRPG compared to AMI and ICSI audio files (see \cref{subsection:diar_result,subsection:manual_labelled_eval}), which we consider to be similar except for the unique TTRPG properties (see \cref{subsection:ami_dataset}).
We found evidence that a low-resource method of annotating a TTRPG dataset does not conceal the fact that a diarizer gets confused by TTRPGs' properties. Additionally, we found that it could be advantageous to annotate whether utterances are in-character or descriptive, to be able to evaluate the diarization performance in more depth.
\texttt{pyannote.audio}'s and \texttt{wespeaker}'s confusion increased for TTRPG audios.
Nevertheless, the relative amount of speakers \pyannote found compared to the amount of speakers that actually are in the audio file did not change.
In case of \wespeaker we found that the number \textit{decreased}, indicating that its clustering algorithm gets confused by people changing their voices frequently (see \cref{subsection:amount_speakers}).

The fact that TTRPG data confuse the diarizers aligns with the findings of other work showing that diarization performance is not yet good enough for other naturalistic dialogues such as in-person role-play dialogues in health care education~\cite{medaramitta2021evaluating}.
We conclude that using a dataset of this kind as a challenge or even train a diarizer on it could make diarizers more robust.

\section{Limitations}

This work presents a proof of concept which needs further investigation to ensure our findings can be generalized. Other SOTA diarizers should be tested on top of \texttt{pyannote.audio} and \texttt{wespeaker}.
Additionally, we only compared the diarization performance to the AMI and ICSI corpora and had a relatively small sample size.
As mentioned in \cref{subsection:weaknesses_reference}, the ground truth for the TTRPGs is imperfect, however as shown in \cref{subsection:manual_labelled_eval} we showed evidence that the imperfect ground truth do not compromise our core findings.
Our claim concerning the reduction in confusion error rates for manually aligned audio in Section 5.4 needs to be further validated statistically. 
This can be done in a larger study that examines multiple sets of manual and automatic annotated audio.

As \texttt{pyannote.audio} has been trained on parts of the AMI dataset, we probably have a bias towards the diarization of AMI, even though we tested \pyannote on the AMI audio files that have not been used to train it.
Nevertheless, our comparison between AMI and TTRPG could be unfair. However, as we also used the ICSI corpus and \wespeaker had similar results about the confusion rate, we have evidence that the bias towards AMI did not influence our results.

We did not examine whether our results would still hold after taking into account of lexical information which has been shown to improve DER over acoustic only systems \cite{Flemotomos2020Lexical}.
Additionally, the acted speech in TTRPG context may not be helpful for training diarizers that are meant to be applied to naturalistic speech, as acted conversation does not represent natural conversation accurately~\cite{schuller2010cross}.

We were not able to take into account the voice-artist skills, and age group (children vs. adults) of the players, since our players were all adults and their voice-artist skills were not quantified. The players’ attributes, such as these, must be considered to fully establish TTRPGs as benchmarks for diarization systems.

\section{Ethical considerations}\label{subsection:ethics}

Our TTRPG dataset was compiled using  publicly available videos and subtitles.
The processing was performed on an off-line computing cluster, meaning we did not upload the speaker files to any third-party.
As this paper is a proof of concept and the data are not published or shared to any third-party, we had no ethical concerns in experimenting with the audio files. %

The findings of our paper and the publishing of the YouTube links to the TTRPG videos we used puts TTRPG content at risk to be downloaded for dataset-creation or used as training data without the creators' consent.
We appeal to researchers to only create datasets and train models with data for which consent of the creators was given.
We believe that we have only slightly increased the risk of YouTube videos being used without the creators' consent, as the videos could already be copied relatively easy from YouTube.

The usage of \texttt{pyannote.audio}, \texttt{wespeaker}, the AMI and the ICSI dataset in this work are compatible with their intended use for research.

The involved university does not require IRB approval for this kind of study, which uses publicly available data.

We do not see any other concrete risks concerning dual use of our research results.
Of course, in the long run, any research results on AI methods based on large language models could potentially be used in contexts of harmful and unsafe applications of AI.
But this danger is rather low in our concrete case.

\bibliography{custom}

\appendix

\section{Mann-Whitney U test results}\label{sect:MWU-appendix}

For a better overview, we provide the values of the Mann-Whitney U tests~\cite{mann1947test} in this paper.

\begin{table*}[ht]
  \centering
  \begin{tabular}{lllllll}
    \textbf{Model} & \textbf{\shortstack{Computational\\property}} & \textbf{Set $1$ (Average)} & \textbf{Set $2$ (Average)} & \textbf{Type} & \textbf{U} & \textbf{p}\\
    \hline
    PA & $\frac{\text{Detected speaker no.}}{\text{Actual speaker no.}}$ & AMI ($1.53$) & ICSI ($1.43$) & two-sided & $684$ & $0.38$\\
    PA & $\frac{\text{Detected speaker no.}}{\text{Actual speaker no.}}$ & AMI ($1.53$) & TTRPG ($1.56$) & two-sided & $155$ & $0.70$\\
    PA & $\frac{\text{Detected speaker no.}}{\text{Actual speaker no.}}$ & ICSI ($1.43$) & TTRPG ($1.56$) & two-sided & $606.5$ & $0.11$\\
    WS & $\frac{\text{Detected speaker no.}}{\text{Actual speaker no.}}$ & AMI ($1.17$) & ICSI ($1.01$) & two-sided & $9458$ & $2 \cdot 10^{-11}$***\\
    WS & $\frac{\text{Detected speaker no.}}{\text{Actual speaker no.}}$ & AMI ($1.17$) & TTRPG ($0.30$) & two-sided & $155$ & $2 \cdot 10^{-15}$***\\
    WS & $\frac{\text{Detected speaker no.}}{\text{Actual speaker no.}}$ & ICSI ($1.01$) & TTRPG ($0.30$) & two-sided & $606.5$ & $5 \cdot 10^{-12}$***\\
    PA/WS & $\frac{\text{Detected speaker no.}}{\text{Actual speaker no.}}$ & AMI PA ($1.53$) & AMI WS ($1.17$) & one-sided & $2169$ & $4 \cdot 10^{-6}$***\\
    PA/WS & $\frac{\text{Detected speaker no.}}{\text{Actual speaker no.}}$ & ICSI PA ($1.43$) & ICSI WS ($1.01$) & one-sided & $5195$ & $6 \cdot 10^{-19}$***
  \end{tabular}
  \caption{
    The Mann-Whitney U test~\cite{mann1947test} results of the tests of \cref{subsection:amount_speakers}. PA: \texttt{pyannote.audio}, WS: \texttt{wespeaker}.
  }\label{tab:mann_whitney_u_amount_speakers}
\end{table*}

\begin{table*}[ht]
  \centering
  \begin{tabular}{lllllll}
    \textbf{Model} & \textbf{\shortstack{Computational\\property}} & \textbf{Set $1$ (Average)} & \textbf{Set $2$ (Average)} & \textbf{Type} & \textbf{U} & \textbf{p}\\
    \hline
    PA & DER & AMI ($0.12$) & TTRPG ($0.33$) & one-sided & $14$ & $1 \cdot 10^{-6}$***\\
    PA & DER & ICSI ($0.28$) & TTRPG ($0.33$) & one-sided & $492$ & $0.004$**\\
    WS & DER & AMI ($0.19$) & TTRPG ($0.48$) & one-sided & $126$ & $1 \cdot 10^{-12}$***\\
    WS & DER & ICSI ($0.27$) & TTRPG ($0.48$) & one-sided & $152$ & $9 \cdot 10^{-9}$***\\
    PA & MD & AMI ($0.066$) & ICSI ($0.018$) & two-sided & $1147$ & $1 \cdot 10^{-8}$***\\
    PA & MD & AMI ($0.066$) & TTRPG ($0.066$) & two-sided & $176$ & $0.82$\\
    PA & MD & ICSI  ($0.018$) & TTRPG ($0.066$) & two-sided & $58$ & $1 \cdot 10^{-10}$***\\
    WS & MD & AMI ($0.13$) & ICSI ($0.052$) & two-sided & $11910$ & $2 \cdot 10^{-28}$***\\
    WS & MD & AMI ($0.13$) & TTRPG ($0.11$) & two-sided & $2079$ & $0.18$\\
    WS & MD & ICSI ($0.052$) & TTRPG ($0.11$) & two-sided & $66$ & $2 \cdot 10^{-10}$***\\
    PA & FA & AMI ($0.018$) & TTRPG ($0.15$) & one-sided & $4$ & $3 \cdot 10^{-7}$***\\
    PA & FA & ICSI ($0.24$) & TTRPG ($0.15$) & two-sided & $1224$ & $1 \cdot 10^{-4}$***\\
    WS & FA & AMI ($0.040$) & TTRPG ($0.078$) & one-sided & $538$ & $1 \cdot 10^{-7}$***\\
    WS & FA & ICSI ($0.21$) & TTRPG ($0.078$) & two-sided & $1473$ & $1 \cdot 10^{-9}$***\\
    PA & Conf & AMI ($0.034$) & ICSI ($0.022$) & two-sided & $700$ & $0.30$\\
    PA & Conf & AMI ($0.034$) & TTRPG ($0.12$) & one-sided & $32$ & $2 \cdot 10^{-5}$***\\
    PA & Conf & ICSI ($0.022$) & TTRPG ($0.12$) & one-sided & $75$ & $1 \cdot 10^{-10}$***\\
    WS & Conf & AMI ($0.022$) & ICSI ($0.012$) & two-sided & $9411$ & $8 \cdot 10^{-10}$***\\
    WS & Conf & AMI ($0.022$) & TTRPG ($0.29$) & one-sided & $165$ & $7 \cdot 10^{-12}$***\\
    WS & Conf & ICSI ($0.012$) & TTRPG ($0.29$) & one-sided & $54$ & $4 \cdot 10^{-11}$***\\
    PA & Conf (SG) & AMI($0.19$) & TTRPG ($0.15$) & two-sided & $207$ & $0.24$\\
    PA & Conf (SG) & ICSI ($0.14$) & TTRPG ($0.15$) & two-sided & $605$ & $0.11$\\
    WS & Conf (SG) & AMI ($0.018$) & TTRPG ($0.29$) & one-sided & $148$ & $4 \cdot 10^{-12}$***\\
    WS & Conf (SG) & ICSI ($0.019$) & TTRPG ($0.29$) & one-sided & $54$ & $3 \cdot 10^{-10}$***
  \end{tabular}
  \caption{
    The Mann-Whitney U test~\cite{mann1947test} results of the tests of \cref{subsection:diar_result}. PA: \texttt{pyannote.audio}, WS: \texttt{wespeaker}, DER: diarization error rate, MD: missed detection rate, FA: false alarm rate, Conf: confusion rate, SG: speaker given.
  }\label{tab:mann_whitney_u_diar_result}
\end{table*}

\clearpage

\section{Additional Figures}\label{sec:fig-appendix}

\begin{figure}[ht]
  \centering
  \pdftooltip{\includegraphics[width=\linewidth]{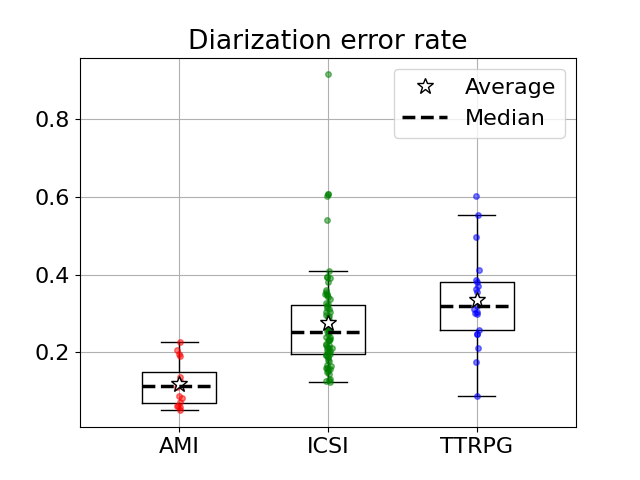}}{A boxplot diagram showing the diarization error rate for AMI, ICSI and TTRPG. AMI: Average 0.12, median 0.11, 1. quartile 0.07, 3. quartile 0.15, interquartile range from 0.05 to 0.23, no outliners. ICSI: Average 0.28, median 0.25, 1. quartile 0.19, 3. quartile 0.32, interquartile range from 0.12 to 0.41, outliners at 0.54, 0.60, 0.61, 0.61 and 0.92. TTRPG: Average 0.33, median 0.32, 1. quartile 0.26, 3. quartile 0.38, interquartile range from 0.09 to 0.55, outliner at 0.60.}
  \caption{The DER by \texttt{pyannote.audio}. A one-sided Mann-Whitney U test~\cite{mann1947test} showed that the TTRPG dataset was significant higher than AMI ($U=14$, $p=1 \cdot 10^{-6}$) or ICSI ($U=492$, $p=0.004$).}
  \label{fig:der_box}
\end{figure}

\begin{figure}[ht]
  \centering
  \pdftooltip{\includegraphics[width=\linewidth]{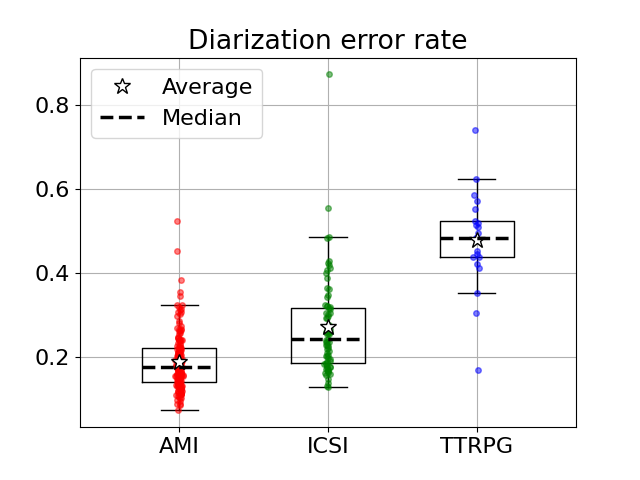}}{A boxplot diagram showing the diarization error rate for AMI, ICSI and TTRPG. AMI: Average 0.188, median 0.176, 1. quartile 0.140, 3. quartile 0.220, interquartile range from 0.072 to 0.346, outliners at 0.35, 0.383, 0.452 and 0.524. ICSI: Average 0.270, median 0.242, 1. quartile 0.187, 3. quartile 0.318, interquartile range from 0.129 to 0.485, outliners at 0.554 and 0.874. TTRPG: Average 0.478, median 0.484, 1. quartile 0.438, 3. quartile 0.523, interquartile range from 0.353 to 0.624, outliner at 0.170, 0.305 and 0.741.}
  \caption{The DER by \texttt{wespeaker}. A one-sided Mann-Whitney U test~\cite{mann1947test} showed that the TTRPG dataset was significant higher than AMI ($U=126$, $p=2 \cdot 10^{-12}$) or ICSI ($U=152$, $p=9 \cdot 10^{-9}$).}
  \label{fig:der_box_ws}
\end{figure}

\begin{figure}[ht]
  \centering
  \pdftooltip{\includegraphics[width=\linewidth]{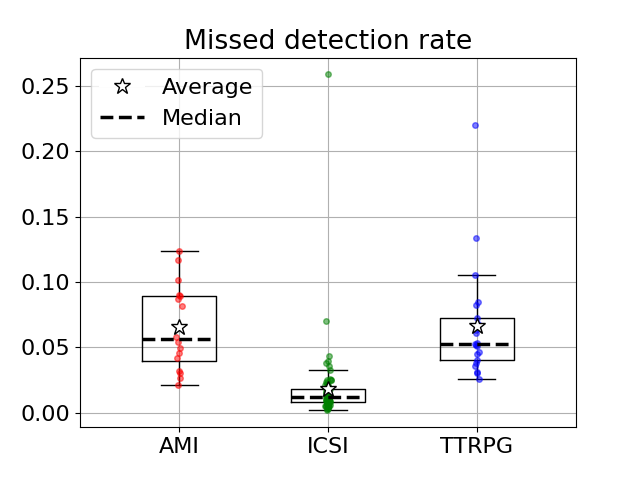}}{A boxplot diagram showing the missed detection rate for AMI, ICSI and TTRPG. AMI: Average 0.066, median 0.056, 1. quartile 0.039, 3. quartile 0.090, interquartile range from 0.021 to 0.124, no outliners. ICSI: Average 0.018, median 0.012, 1. quartile 0.008, 3. quartile 0.018, interquartile range from 0.002 to 0.032, outliners at 0.036, 0.038, 0.040, 0.043, 0.070 and 0.26. TTRPG: Average 0.066, median 0.052, 1. quartile 0.040, 3. quartile 0.072, interquartile range from 0.026 to 0.106, outliners at 0.13 and 0.22.}
  \caption{The missed detection by \texttt{pyannote.audio}. A two-sided Mann-Whitney U test~\cite{mann1947test} did not show significant differences between AMI and TTRPG ($U=176$, $p=0.8$), but ICSI differs significantly from AMI ($U=1147$, $p=1 \cdot 10^{-8}$) and TTRPG ($U=58$, $p=1 \cdot 10^{-10}$).}
  \label{fig:missed_detection_box}
\end{figure}

\begin{figure}[ht]
  \centering
  \pdftooltip{\includegraphics[width=\linewidth]{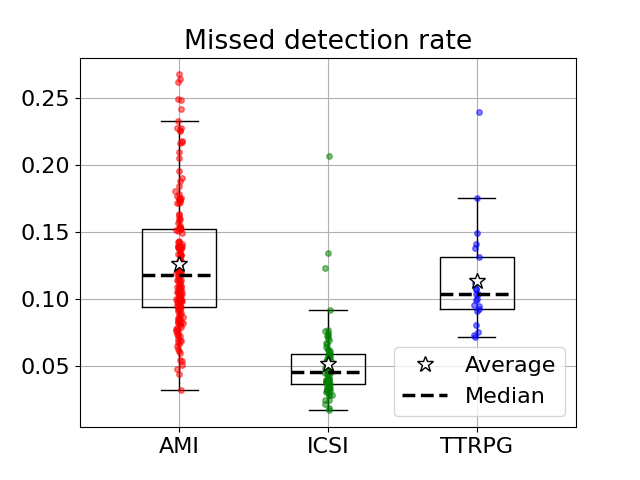}}{A boxplot diagram showing the missed detection rate for AMI, ICSI and TTRPG. AMI: Average 0.126, median 0.118, 1. quartile 0.094, 3. quartile 0.152, interquartile range from 0.032 to 0.232, outliners at 0.242, 0.248, 0.249, 0.264, 0.261 and 0.267. ICSI: Average 0.052, median 0.046, 1. quartile 0.037, 3. quartile 0.059, interquartile range from 0.017 to 0.092, outliners at 0.124, 0.134 and 0.207. TTRPG: Average 0.114, median 0.104, 1. quartile 0.092, 3. quartile 0.131, interquartile range from 0.072 to 0.175, outliner at 0.240.}
  \caption{The missed detection by \texttt{wespeaker}. A two-sided Mann-Whitney U test~\cite{mann1947test} did not show significant differences between AMI and TTRPG ($U=2079$, $p=0.2$), but ICSI differs significantly from AMI ($U=11910$, $p=2 \cdot 10^{-28}$) and TTRPG ($U=66$, $p=2 \cdot 10^{-10}$).}
  \label{fig:missed_detection_box_ws}
\end{figure}

\begin{figure}[ht]
  \centering
  \pdftooltip{\includegraphics[width=\linewidth]{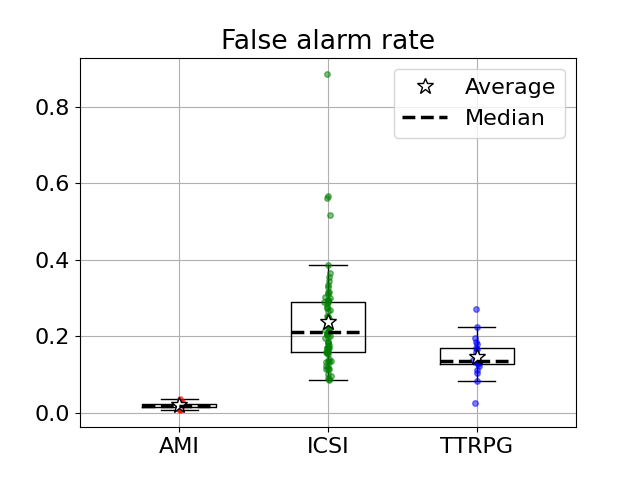}}{A boxplot diagram showing the false alarm rate for AMI, ICSI and TTRPG. AMI: Average 0.018, median 0.018, 1. quartile 0.015, 3. quartile 0.023, interquartile range from 0.006 to 0.035, no outliners. ICSI: Average 0.24, median 0.21, 1. quartile 0.16, 3. quartile 0.29, interquartile range from 0.086 to 0.38, outliners at 0.52, 0.56 and 0.88. TTRPG: Average 0.145, median 0.134, 1. quartile 0.126, 3. quartile 0.168, interquartile range from 0.083 to 0.224, outliners at 0.025 and 0.271.}
  \caption{The false alarm by \texttt{pyannote.audio}. A two-sided Mann-Whitney U test~\cite{mann1947test} showed a significant difference between ICSI and TTRPG ($U=1224$, $p=1 \cdot 10^{-4}$). A one-sided test showed the TTRPG rates to be significantly higher than the AMI rates ($U=4$, $p=3 \cdot 10^{-7}$).}
  \label{fig:false_alarm_box}
\end{figure}

\begin{figure}[ht]
  \centering
  \pdftooltip{\includegraphics[width=\linewidth]{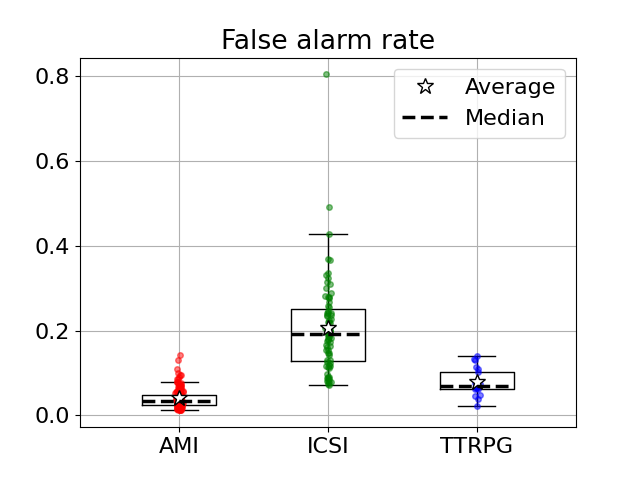}}{A boxplot diagram showing the false alarm rate for AMI, ICSI and TTRPG. AMI: Average 0.040, median 0.034, 1. quartile 0.025, 3. quartile 0.047, interquartile range from 0.011 to 0.086, outliners at 0.087, 0.089, 0.094, 0.096, 0.099, 0.110, 0.130 and 0.142. ICSI: Average 0.207, median 0.192, 1. quartile 0.128, 3. quartile 0.251, interquartile range from 0.073 to 0.428, outliners at 0.491 and 0.804. TTRPG: Average 0.078, median 0.069, 1. quartile 0.062, 3. quartile 0.102, interquartile range from 0.022 to 0.139, no outliners.}
  \caption{The false alarm by \texttt{wespeaker}. A two-sided Mann-Whitney U test~\cite{mann1947test} showed a significant difference between ICSI and TTRPG ($U=1473$, $p=1 \cdot 10^{-9}$). A one-sided test showed the TTRPG rates to be significantly higher than the AMI rates ($U=538$, $p=1 \cdot 10^{-7}$).}
  \label{fig:false_alarm_box_ws}
\end{figure}

\begin{figure}[ht]
  \centering
  \pdftooltip{\includegraphics[width=\linewidth]{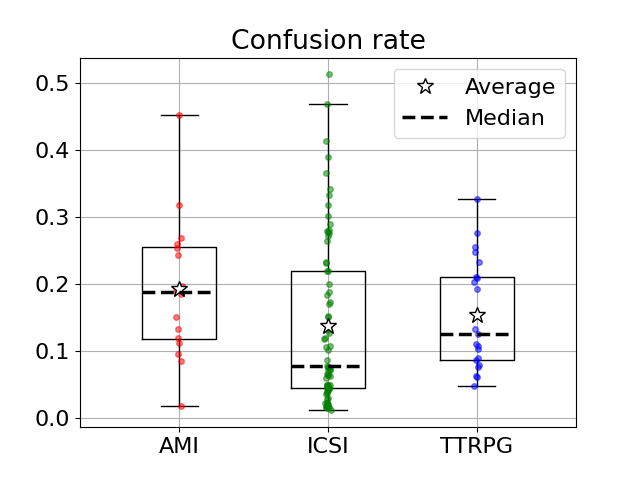}}{A boxplot diagram showing the confusion rate for AMI, ICSI and TTRPG. AMI: Average 0.193, median 0.188, 1. quartile 0.118, 3. quartile 0.256, interquartile range from 0.017 to 0.452, no outliners. ICSI: Average 0.138, median 0.077, 1. quartile 0.043, 3. quartile 0.219, interquartile range from 0.011 to 0.469, outliner at 0.513. TTRPG: Average 0.154, median 0.126, 1. quartile 0.086, 3. quartile 0.211, interquartile range from 0.048 to 0.328, no outliners.}
  \caption{The confusion by \texttt{pyannote.audio} if the numbers of speakers has been given. A two-sided Mann-Whitney U test~\cite{mann1947test} did not show significant differences between AMI and TTRPG ($U=207$, $p=0.2$) or ICSI and TTRPG ($U=605$, $p=0.1$).}
  \label{fig:confusion_speaker_box}
\end{figure}

\begin{figure}[ht]
  \centering
  \pdftooltip{\includegraphics[width=\linewidth]{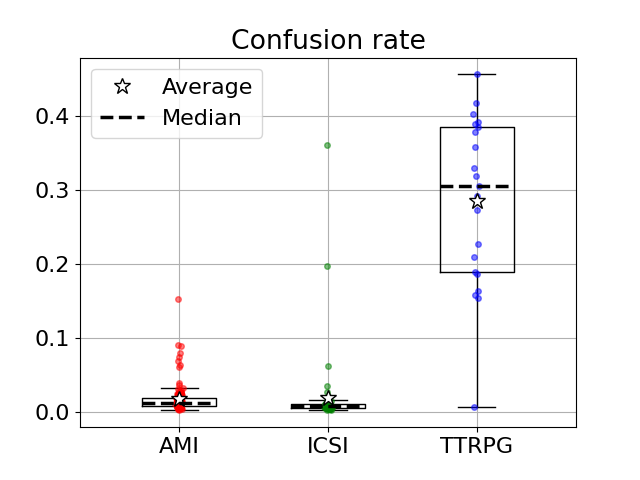}}{A boxplot diagram showing the confusion rate for AMI, ICSI and TTRPG. AMI: Average 0.018, median 0.013, 1. quartile 0.009, 3. quartile 0.019, interquartile range from 0.003 to 0.037, outliners at 0.040, 0.061, 0.064, 0.070, 0.074, 0.080, 0.089, 0.091 and 0.153. ICSI: Average 0.0.019, median 0.008, 1. quartile 0.006, 3. quartile 0.011, interquartile range from 0.002 to 0.022, outliners at 0.028, 0.036, 0.062, 0.20 and 0.361. TTRPG: Average 0.285, median 0.305, 1. quartile 0.189, 3. quartile 0.385, interquartile range from 0.007 to 0.456, no outliners.}
  \caption{The confusion by \texttt{wespeaker} if the number of speakers has been given. A two-sided Mann-Whitney U test~\cite{mann1947test} showed significant differences between AMI and TTRPG ($U=148$, $p=1 \cdot 10^{-12}$) or ICSI and TTRPG ($U=54$, $p=3 \cdot 10^{-10}$).}
  \label{fig:confusion_box_speaker_ws}
\end{figure}

\end{document}